%% file: root.tex
\documentclass[letterpaper, 10 pt, journal, twoside]{IEEEtran}

\usepackage{graphicx}
\usepackage{amsmath}
\usepackage{amssymb}
\usepackage{tabularx}
\usepackage{multirow}
\usepackage[dvipsnames]{xcolor}
\usepackage{comment}
\usepackage{epsfig}
\usepackage{lineno} 

\begin{document}
\title{\LARGE \bf
Learn to Predict How Humans Manipulate Large-sized Objects \\ from Interactive Motions
}

\author{Weilin Wan$^{1}$, Lei Yang$^{1,4}$, Lingjie Liu$^{2}$, Zhuoying Zhang$^{1}$, Ruixing Jia$^{1}$, Yi-King Choi$^{1,4}$, \\ Jia Pan$^{1}$, Christian Theobalt$^{2}$, Taku Komura$^{1}$ and Wenping Wang$^{3}$
\thanks{Manuscript received: September 9, 2021; Accepted: November 22, 2021. {\tt\small }}%
\thanks{This paper was recommended for publication by
Editor Angelika Peer upon evaluation of the Associate Editor and Reviewers’
comments.{\tt\small }}%
\thanks{$^{1}$ Department of Computer Science, The University of Hong Kong, Hong Kong SAR
        {\tt\small }}%
\thanks{$^{2}$ Max-Planck-Institute for Informatics, Saarbruecken, Germany {\tt\small }}%
\thanks{C. Theobalt was supported by the ERC Consolidator Grant 4DRepLy (770784). L. Liu was supported by Lise Meitner Postdoctoral Fellowship. {\tt\small }}%
\thanks{$^{3}$ Department of Computer Science and Engineering, Texas A\&M University, TX, USA
        {\tt\small }}%
\thanks{$^{4}$ Centre for Garment Production Limited, Hong Kong SAR
        {\tt\small }}%
\thanks{        
This work was partially supported by the Innovation and Technology Commission of the HKSAR Goverment under the InnoHK initiative.{\tt\small }}%
\thanks{        
Digital Object
Identifier (DOI): see top of this page.{\tt\small }}%
}


\maketitle
\input{pages/abstract.tex}
\begin{IEEEkeywords}
Intention Recognition, Human-Robot Collaboration, Datasets for Human Motion
\end{IEEEkeywords}
\input{pages/intro.tex}

\input{pages/relatedWork.tex}

\input{pages/methodology.tex}

\input{pages/dataset.tex}

\input{pages/Experiments}

\input{pages/demo}

\input{pages/conclusion}

{\small
\bibliographystyle{IEEEtranS}
\bibliography{references}
}

\end{document}

%% file: pages/abstract.tex
\begin{abstract}

Understanding human intentions during interactions has been a long-lasting theme, that has applications in human-robot interaction, virtual reality and surveillance. In this study, we focus on full-body human interactions with large-sized daily objects and aim to predict the future states of objects and humans given a sequential observation of human-object interaction. As there is no such dataset dedicated to full-body human interactions with large-sized daily objects, we collected a large-scale dataset containing thousands of interactions for training and evaluation purposes. We also observe that an object's intrinsic physical properties are useful for the object motion prediction, and thus design a set of object dynamic descriptors to encode such intrinsic properties. 
We treat the object dynamic descriptors as a new modality and propose a graph neural network, HO-GCN, to fuse motion data and dynamic descriptors for the prediction task.
We show the proposed network that consumes dynamic descriptors can achieve state-of-the-art prediction results and help the network better generalize to unseen objects. We also demonstrate the predicted results are useful for human-robot collaborations.
    
\end{abstract}

%% file: pages/intro.tex
\section{Introduction}\label{sec:intro}

Predicting human intentions to manipulate or carry objects, or more specifically, the future states of such interactions from a given sequential observation, has applications in robotic prosthesis and human-robot interaction~\cite{cheng2019purposive,xia2018gibson,lafleche2018robot_Toronto_Robot,koppula2013anticipating_anticipate,koppula2013learning_earlyHO}, where the accuracy of such prediction can strongly affect the embodied perception and VR/3D gaming~\cite{kasahara2017malleable}, where compensation of latency through such prediction is needed for improving the user experience. 
Despite the high demand, research for such prediction has been limited to small-sized objects~\cite{taheri2020grab,corona2020context} or complex scenes~\cite{cao2020long,zhangplace,Mandery2015a}. Research for interactions with larger objects, such as chairs or boxes, that involve full-body motions, is rather under-explored and requires considering physics for reliable prediction. 

Learning the physical interactions between the human body and arbitrary objects is hard due to the large variation of object geometries, the complex physics involved, and a large amount of training needed~\cite{finn2016unsupervised,battaglia2016interaction,janner2018reasoning,ehsani2020use_the_force}.  Ehsani et al.~\cite{ehsani2020use_the_force} propose to integrate physics into a neural network and leverage it for object motion prediction. Nevertheless, the exact geometry of the object is required for the neural framework. They also require conducting physical simulation within the training loop, which is expensive and time-consuming for good generalization. 

\begin{figure}
    \centering
    \includegraphics[width=0.49\textwidth]{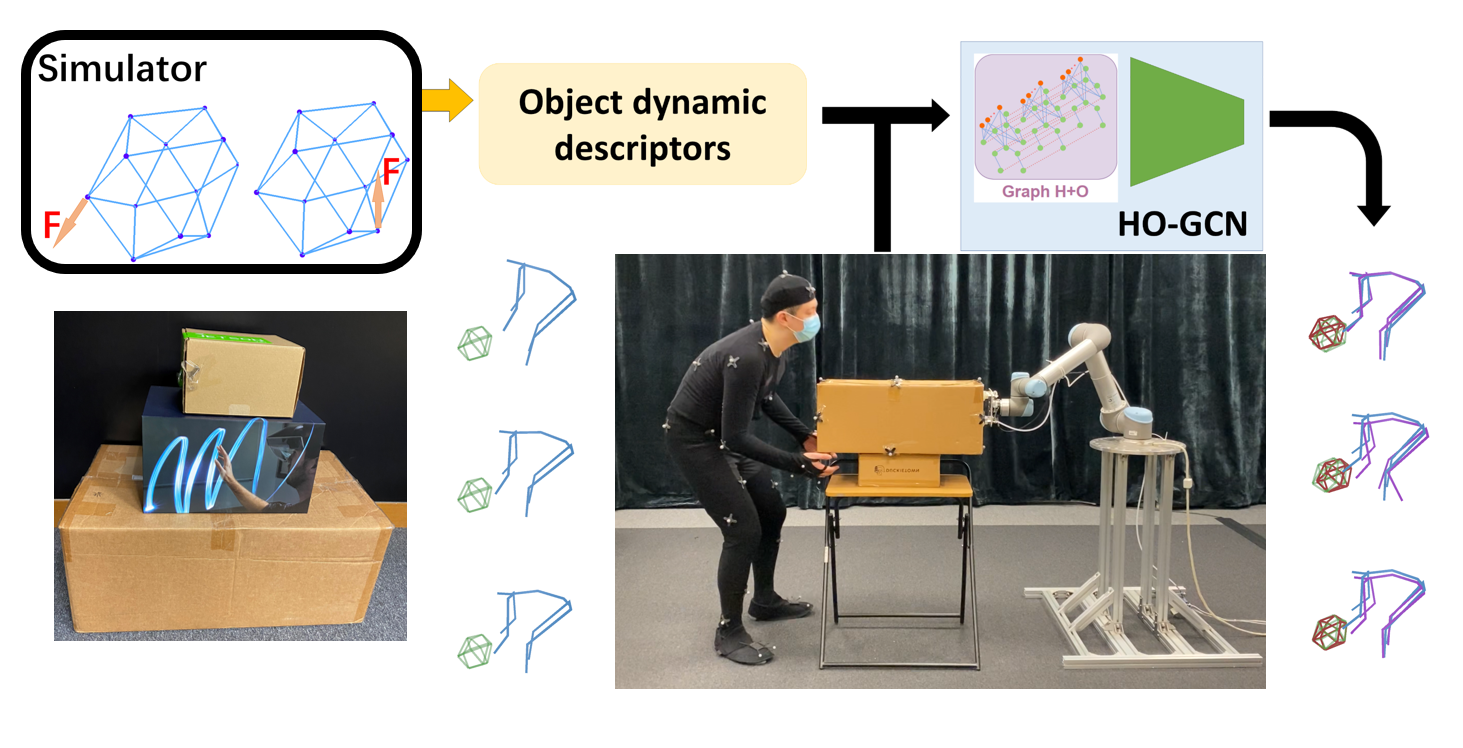}
    \caption{
    We introduce a large-scale dataset on full-body human interactions with large-sized objects, such as chairs and boxes, using a motion capture system. Our goal is to predict the human-object motion at future time steps (where predicted human and object are in purple and red, respectively). To this end, we propose to leverage the \textit{object dynamic descriptors} and design a neural network, \textit{HO-GCN}, to fuse data of different modalities. We also showcase a human-robot collaborative task to validate the proposed method. }
    \label{fig:teaser}
\end{figure}

We make two efforts to address the aforementioned challenges. First, in response to the lack of datasets concerning interactions between humans and large-sized objects, we construct such a dataset. It contains 384K frames from 508 full-body motion capture (MoCap) videos, involving 12 daily objects from 6 categories and different actions. Not only included are the human body motions but also the 6 degrees-of-freedom (DOF) motions of the objects that are represented by 12 keypoints. The interaction take place in diverse contexts, such as transporting the objects forward or backward, rotating the objects, or more complex compositional motions (e.g., lift and carry forward), depending on the affordances and functionality of the object. The data were fully preprocessed and will be available to the community upon the publication of the paper. To the best of our knowledge, this is the first large-scale dataset focusing full-body interactions with large-sized objects.

To avoid frequent calls for physical simulators during training, we propose a novel descriptor that encodes the object's intrinsic dynamics obtained from simulations. In particular, a conceptual model for each category of objects (e.g., chairs) is constructed, which is represented by a set of keypoints abstracting the general shape of this object category. Then, we simulate its responses under forces as a rigid-body system and acquire the intrinsic dynamic properties. Thus, the \textit{object dynamic descriptors} are defined as how the keypoint-based conceptual model is transformed under given forces. 
As shown in our experiments, our method using the object dynamic descriptors can achieve state-of-the-art performances regarding the prediction of object motions even when the object instance is unseen in the training set. 
For human-robot collaborative tasks like lifting a box or handing over an object, this is crucial as reliable prediction of the object motions can enable the robot to provide desirable assistive operations to the human partners.

We also design a novel graph convolutional neural network (HO-GCN) that takes as input the human-object interactive motions as well as the object dynamic descriptors as shown in Fig.~\ref{fig:teaser}. 
We adopt the spatial-temporal graph convolution proposed in \cite{yan2018spatial_ST_GCN} to learn motion features and predict the future interactive motions based on the input sequence. 
We evaluate the proposed method as well as several state-of-the-art methods on our collected dataset. We also showcase that the predicted interactive motion can enable robot assistance in labor-intensive tasks, such as transporting a box.

Our technical contributions are threefold:
\begin{enumerate}
    \item We contribute the \textit{first} large-scale dataset concerning human full-body interactions with large-sized daily objects;
    \item We propose to consider the object's intrinsic dynamics as an extra modality for enhancing prediction of the object future poses during human-object interactions; 
    \item We design a novel graph convolutional neural network that fuses the observed human-object interaction sequence and the object intrinsic dynamics for the prediction task.
\end{enumerate}

%% file: pages/relatedWork.tex
\section{Related Work}\label{sec:related_work}

\subsection{Learning from human skeleton data}
Recognizing human activities from skeletal data receives increasing research attention as the acquisition of human skeletons becomes easier, such as employing motion capture techniques. With the development of deep learning, many attempts are made to learn features from the temporal sequences of the human skeleton data for action recognition or motion prediction. 
These works can be roughly categorized into RNN/LSTM-based methods which recursively process the temporal information \cite{si2018skeleton_rnn3, zhang2017view_rnn1, lee2017ensemble_rnn2,corona2020context}, and CNN-based methods which maps the temporal information into hidden features using convolution \cite{tang2018deep_cnn1, du2015skeleton_cnn4, li2018co_cnn2, ding2017investigation_cnn3,butepage2017deep}. 

Since a human skeleton is naturally a graph structure, Graph Convolutional Networks (GCNs) are proposed for processing human skeleton data as well. 
Yan et al.~\cite{yan2018spatial_ST_GCN} propose spatial-temporal GCN to efficiently extract the spatial and temporal features from the skeleton inputs for action recognition. Li et al.~\cite{li2019actional_links} extend the human skeleton graph in GCN with extra links to capture more dependencies and explore significant features of movements. Even GCNs have shown to be effective in representing human skeletons as graphs, only a few attempts have been made to couple human skeletons and objects in GCNs. Kim et al.~\cite{kim2019skeleton_korea} incorporate a single point containing objects' positional information to the human skeleton graph for classifying human actions. Cui et al.~\cite{cui2020learning} propose to learn additional connectivity among joints besides their natural linkages for motion prediction. 

In our work, we design HO-GCN, a novel architecture that fuses object information and human skeleton information to predict the 6DOF motion of the object. We compare our proposed method with C-TE~\cite{butepage2017deep} and CAHMP~\cite{corona2020context}. The former is a convolution-based method using a mirrored encoder-decoder framework to process the temporal inputs and generate predicted skeleton motion sequences. The latter is a recently proposed RNN-based method for predicting interactive motions between humans and the static environments or smaller objects.

\subsection{Human-object interaction dataset}

Existing 3D human datasets are mainly designed for action recognition or human motion prediction tasks \cite{h36m_pami, Liu_2019_NTURGBD120, Shahroudy_2016_NTURGBD, xia2012view_data, ofli2013berkeley_data}, which contain a limited number of human-object interaction cases and do not provide 3D ground truth position of objects. There have also been works focusing on hand-object interactions. Although these datasets usually provide sufficient spatial information of the objects, they focus on hand-held objects that do not involve full-body movements in the sequences. Ehsani et al.~\cite{ehsani2020use_the_force} propose a dataset with 174 RGB hand-object interaction videos and 6D object pose annotation. Grasping Actions with Body (GRAB) \cite{taheri2020grab} is a large-scale dataset focusing on human grasping actions, which presents 1334 videos with body information and object 3D models. First-Person Hand Action (FPHA) \cite{garcia2018first_FPHA} provides 1175 RGB videos with hand positions and 6D object poses annotated. 

Whole-Body Human Motion Database \cite{Mandery2015a,Mandery2016a} includes a number of samples including human interactions with the scene context involving tables, cups, ladders, etc. However, the number of direct human operations on these large objects (e.g., the table) is limited.
This motivates us to propose a dataset for understanding full-body interactions with large-sized daily objects such as tables, chairs, and boxes. In these scenarios, human poses need to be adapted to the object properties, such as shape or center of mass, making it different from previous works that only model interactions with small  or thin objects (e.g., golf clubs in \cite{li2019estimating}). We also anticipate increasing attention from the community to use such a dataset for learning a prior of human-object joint motions or predicting motions for human-robot interactions.

\subsection{Predicting human intention in human-robot cooperation}
A bulk of literature has contributed to predicting human intentions or motions for efficient and safe human-robot collaborations. 
An early work~\cite{table_lift} proposes a method based on the Gaussian mixed model to learn from human demonstrations and thus adjust the robot's role in the human-robot table lifting task. 
To predict intentions, Hidden Markov Models are widely used (e.g.,~\cite{hr_learn_from_hh_ICRA,lafleche2018robot_Toronto_Robot} for recognizing human movements or interaction selection). Zhao et al.~\cite{collaborative_LSTM_ZhaoXuan} proposed a recurrent network for imitation learning to accomplish object handover tasks between a human and a robot. In this paper, we focus on predicting interactive motions as a human manipulating a large-sized object, and showcase that the output of the proposed prediction method can be leveraged for human-robot collaboration.


%% file: pages/methodology.tex
\section{Methodology}\label{sec:method}

\subsection{Notations and problem formulation}\label{sec:problem_formulation}

The inputs to our prediction task are (1) a sequence of 3D human skeletons $\{\mathbf{X}_{t}\}_{t=0}^{K}$, (2) a sequence of 3D positional information of the object keypoints $\{\mathbf{P}_{t}\}_{t=0}^{K}$, and (3) the object dynamic descriptors $\mathbf{O}$. 
The expected outputs are the 6DoF pose changes of the object $\{\Delta_t\}_{t=K}^{T}$ and the human skeleton motion $\{\mathbf{X}_{t}\}_{t=K}^{T}$.
In what follows, we first detail the definition and acquisition of the object dynamic descriptors $\mathbf{O}$ defined for objects and then present the network design. 

\subsection{Object representation and dynamic descriptors}\label{sec:object_dynamic_desc}

\begin{figure}
    \centering
    \includegraphics[width=0.45\textwidth]{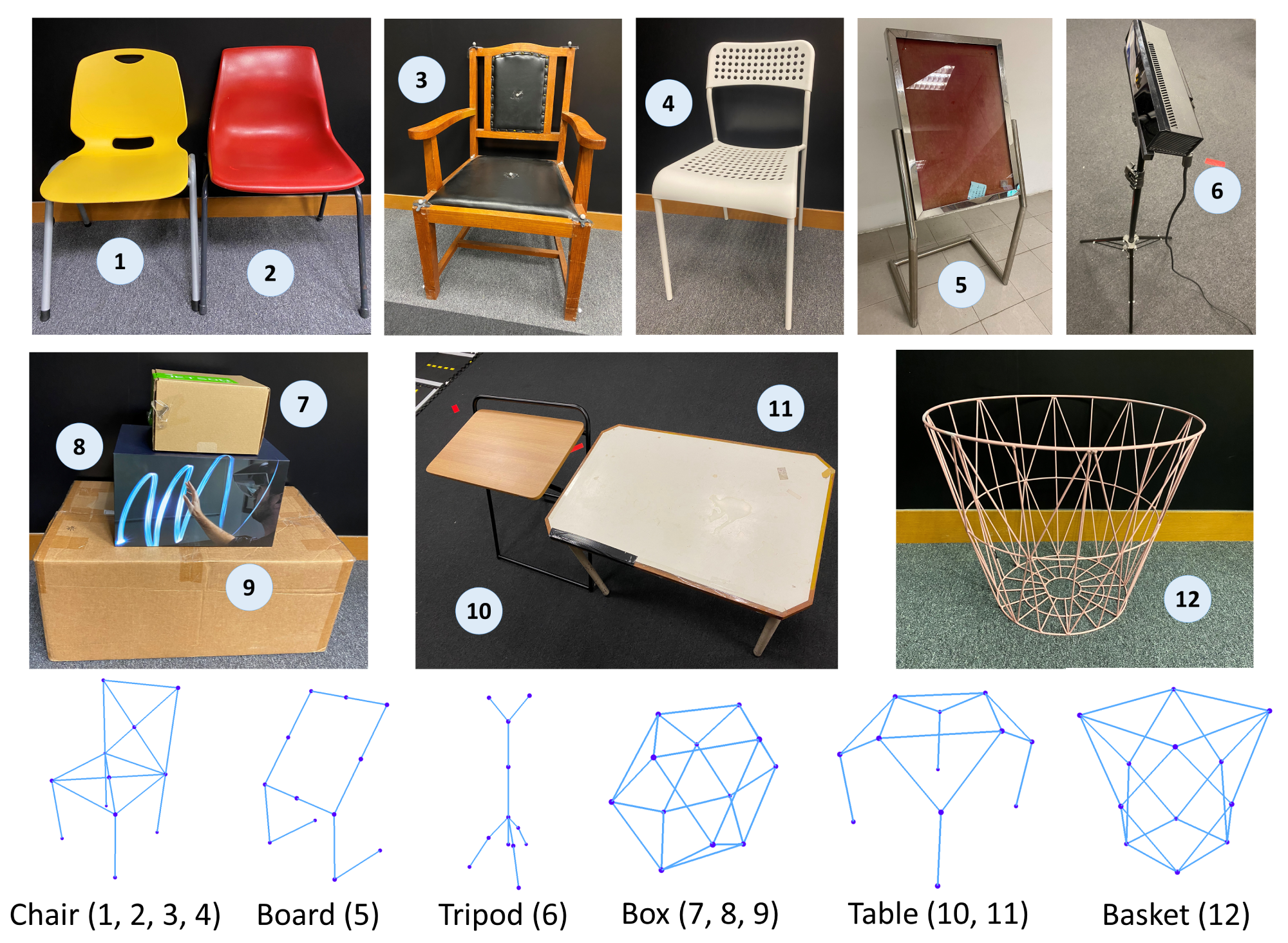}
    \caption{The object instances in our dataset are shown in the top two rows with Object IDs next to them. The conceptual models used in the simulation are shown in the bottom row. Each conceptual model corresponds to an object class. They are modeled as rigid-body systems, and the drawn linkage is for illustration purposes only.}
    \label{fig:objects}
\end{figure}

\textbf{Geometric representation. } 
As many objects from the same class share high similarity in the overall geometry, we assume that objects from the same class can be abstracted by the same geometric conceptual model, and share common object dynamics.
The conceptualized model is defined as a set of sparse keypoints that delineate the geometry of the object, $\mathbf{P} = \{\mathbf{p}_m \in \mathbb{R}^3 \}_{m=1}^M$. 
The conceptualized models of all object categories from the dataset are shown in the bottom row of Fig.~\ref{fig:objects}.
The object dynamic descriptors are defined on the set of the keypoints of the conceptual model.
We assume, for convenience, all conceptual models have the same number of keypoints and thus the same dimensionality of the dynamic descriptors. This allows training on all objects we collected in our dataset.

\begin{figure*}[h!]
\begin{center}
\includegraphics[width=0.9\linewidth, trim=0 140 0 0, clip]{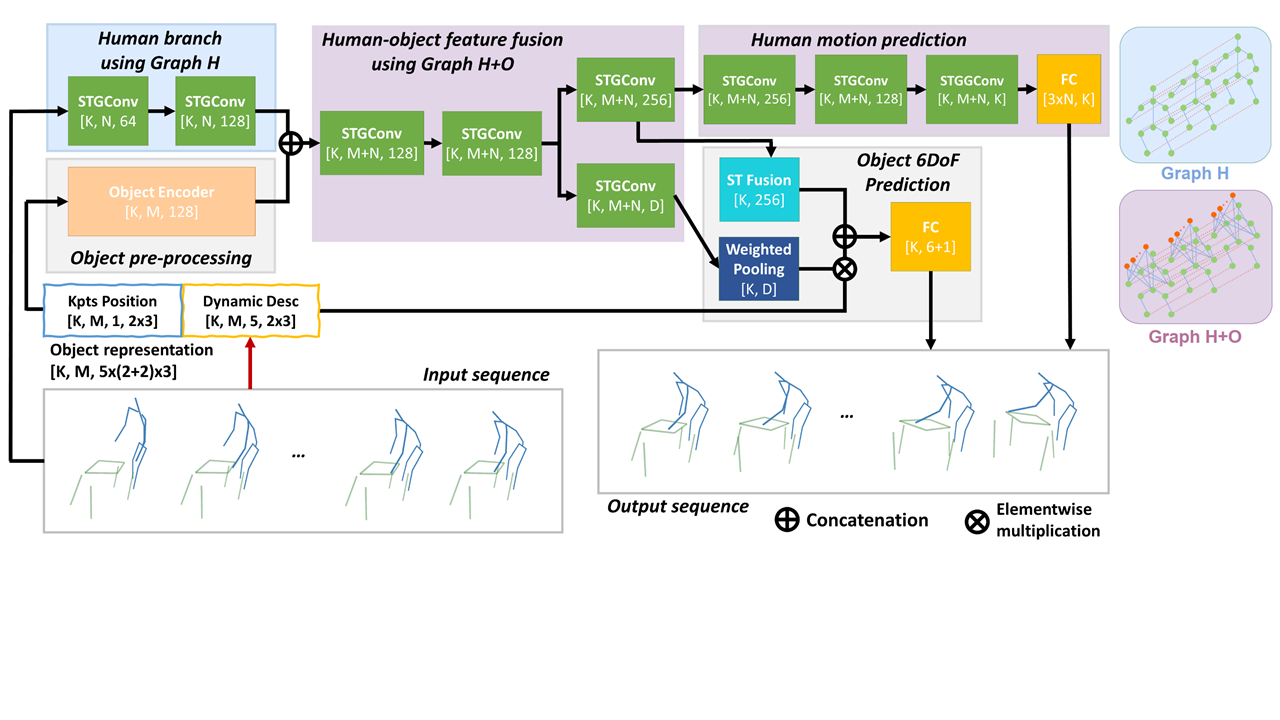}
\end{center}
   \caption{The architecture of HO-GCN.  The network consists of five major blocks. The two parallel blocks at the left end pre-process the human skeleton motion and the object information, and extract respective motion features. The central block using Graph \textit{H+O} fuses the extracted features from the human and object, and performs graph convolutions to achieve global understanding of the input sequence. The upper and lower blocks at the right side respectively predict the human skeletons and the 6DoF changes of the objects in the future frames. \textit{STGConv} and \textit{FC} refers to spatial-temporal graph convolution and fully-connected layers, respectively. D equals $M\times5\times6 $ in our case, where there are 5 candidates of discrete directions.}
\label{fig:hogcn layout}
\end{figure*}

\textbf{Object dynamic descriptors. } 
Object dynamics are a set of intrinsic physical properties that reflect the resulting object motions caused by external forces. In this study, we use $\Delta = g(F)$ to describe how the given unit force $F$ will cause the pose change $\Delta$ of the (conceptual) object.
To reduce the sampling space for the shape and the force, we consider the object as a rigid-body system. Thus, we can perform physical simulations to obtain the relationship between object motions and the forces.
Since forces form a vector space and are linear to their resultant motion, a set of suitable bases of the force space shall suffice to describe this intrinsic dynamic relationship. 

We define a set of unit forces along a candidate set $\mathbb{D}$ of discrete directions, i.e. forward, backward, left, right and up, in the local frame of the object. 
A force applied to keypoint $\mathbf{p}_m$ is denoted $F^j_{m}$ where $j$ indicates a direction from the candidate set.
Then, we collect the resulting pose changes $\Delta^j_m$ subject to the applied force $F^j_m$ from simulations to describe the object dynamics. 
During each simulation, the force is applied to the object within an infinitesimal time interval and the object is modeled as a rigid-body system with a pre-defined floor constraint and the gravity.
Finally, the proposed dynamic descriptor for an object is obtained as $U = \{ \Delta^j_m| m=1...M,~j \in \mathbb{D} \}$ so that it can be fed to common neural networks that consume this semi-structured, discrete information as input.

\subsection{Network structure}\label{sec:network}

To achieve the described task, we propose a novel human-object graph convolutional neural network (HO-GCN) which consists of five major branches as shown in Figure~\ref{fig:hogcn layout}. The \textit{human branch} (top left) is designed for local feature learning from the spatial-temporal graph $H$.
The spatial dimension of graph $H$ is a human skeleton that has $N$ nodes representing human body parts. The temporal dimension of the graph is formed by connecting each node of a human skeleton to its counterpart in temporally adjacent skeletons. Specifically, this branch processes human skeleton data by applying to $H$ the spatial-temporal graph convolutions (denoted \textit{STGConv}) using the distance partitioning strategy~\cite{yan2018spatial_ST_GCN}.

The \textit{object pre-processing branch} aims to fuse the dynamic descriptors and the object keypoints. Specifically, we represent an object as two parts: 1) a $D$-dim vector $U$ 
that represents the dynamic descriptors and 2) 3D positional differences between each object keypoint and both hands in the corresponding input frames. 
For better mining the relations among keypoint positions and descriptors, we broadcast and concatenate the keypoint coordinates with the dynamic descriptors and further process them using a convolution-based object encoder to obtain the object motion features.

The central \textit{human-object fusion branch} then takes the concatenation of human and object motion features from previous branches, and applies spatial-temporal graph convolutions to them based on graph $H+O$. In this graph structure, 
we connect the object keypoints to three joints: left and right hands as well as the hip of a human body, as the movements of the hand and hip joints provides prominent hints related to human-object contact and the proximity between the human and the object for predicting the human's intention. This branch then outputs two tensors for predicting the human and object motions, respectively.

For predicting the 6DoF pose changes of the object in the future frames, we feed the output tensor with $D$ channels from the central $H+O$ branch to the \textit{weighted pooling} module (as shown in Figure~\ref{fig:hogcn layout}). This module is designed for scaling the dynamic descriptors, which are derived by applying unit forces to objects in a simulator. It maps the input tensor to the weighting factors which are then multiplied with the dynamic descriptor vector.

We also employ a spatial-temporal fusion module (\textit{ST Fusion}) to recombine the respective information in spatial and temporal dimensions. This module yields an ST-fused feature containing information about the human-object interaction and is concatenated with the weighted dynamic descriptor vector. 
The concatenated features are then fed into a fully-connected (FC) layer for 6DoF regression. We additionally require the FC layer to regress a probability score $c$ indicating if the object is in motion ($c=1$) and vice versa ($c=0$).

The \textit{human motion prediction branch} performs a series of spatial-temporal graph convolutions based on graph $H+O$ and eventually reduces the channel size of the graph convolution tensor to the number of $K$ output frames. 
Finally, each of the $K$ channels of the tensor is transformed (via a FC layer) to generate $K$ human poses for $K$ output frames, respectively.

\textbf{Loss function. }
The loss function is formulated as:
\begin{align}
    L&=\lambda_1 (\sum_t  \|\widehat{c}_t - c_t\|_2 +  \|\widehat{\Delta}_t - \Delta_t\|_2 + \sum_{p_i \in \mathbf{P}_t}\|\widehat{p}_i - p_i\|_2) \\ \nonumber
    &+ \lambda_2 \sum_{x_i \in \mathbf{X}_t}\|\widehat{x}_i - x_i\|_2,
\end{align}
where notations with a hat, e.g., $\widehat{x}$, denote the predicted value, and those without a hat stand for the ground-truth; $c=\{1, 0\}$ indicates the object in motion or not; $\Delta$ is the pose change; $p_i$ and $x_i$ are the object keypoints and human body joints.
We set the balance coefficients $\lambda_{i=\{1,2\}}$ to 1.0 and 0.5, respectively, for training our network.

%% file: pages/dataset.tex
\section{Human-Object Interaction Dataset}

\begin{table}[]
    \centering
    \resizebox{0.9\linewidth}{!}{
    \begin{tabular}{c}
        \hline
        \textbf{Translation} \\
        Push forward; Pull backward; Move to the left / right (154) \\
        \hline
        \textbf{Rotation} \\
        Out-of-plane: Tilt to left / right / front / back sides (90) \\
        In-plane: Rotate clockwise / anticlockwise (90) \\
        \hline
        \textbf{Compositional actions} \\
        Lift and carry Forward (71)\\
        Lift and place to left/right (60)\\
        Lift and rotate up/down (43) \\
        \hline
    \end{tabular}}
    \caption{Human-object interactive motions in our collected dataset. Numbers in parenthesis indicate the number of video samples.}
    \label{tab:actions}
\end{table}

\textbf{Data collection. }
We collected a dataset of 508 human-object interaction videos, 384K frames in total recorded by OptiTrack~\cite{optitrack}, a motion capture system. 
The average length of the recorded motions in our dataset is $6.3s$ with a standard deviation of $1.4s$. 
Observing how humans may interact the large-size objects in the daily life, we design a set of interactive actions between humans and objects; see Tab.~\ref{tab:actions}. Six actors of different shapes, including two females and four males, participated in data collection process.

Fig.~\ref{fig:objects} shows the 12 large-size objects frequently seen in daily living, i.e., four chairs, a standing board, a tripod, three boxes of different sizes, two tables and a basket. Each of the objects is represented by a geometric abstraction of 12 keypoints delineating its shape, also shown in Fig.~\ref{fig:objects}. We found 12 keypoints are sufficient to describe the geometry of the objects without too much computational overhead or memory consumption.
We follow the keypoint configuration of the conceptual model to attach twelve reflective markers on the real-world objects for tracking their rigid motions.  

\textbf{Data processing. }
We used the OptiTrack motion capture system (at a frame-rate of 120 Hz) to track the human-object interactions. Each MoCap video recorded a complete process of an actor performing given actions and captured the motion data of the human body skeleton and the pose of the target object at each frame.

We use a sliding window of 240 frames and a step size of 12 frames to extract sequences from a given MoCap video. Thus, each extracted sequence has 20 frames equivalent to a 2-sec motion. We use the first ten frames (1 sec) as input and predict human-object interaction in the last ten frames (the other 1 sec). 
We always ensure that the object at the $K$-th frame ($K=10$) in the input sequence to have a small motion within a threshold.
For prediction, we label each frame in the extracted sequence with the 6DOF pose change $\Delta_{gt}$ of the object between two frames:
\begin{equation}
    {\Delta}_{gt}(\delta) = {P}(K+\delta) - {P}(K),
\end{equation}
where ${P}(t)$ is the recorded 6DoF pose of the object at frame $t$. If $||\Delta_{gt}(\delta)|| = 0$, we assign frame $K+\delta$ with the stationary label ($c = 0$) to indicate that the human is yet to move the object; otherwise $c=1$.

\textbf{Data split. }
To test the model generalizability to unseen instances, we reserved all 471 samples of Chairs 3 and 4 (see Fig.~\ref{fig:objects}) to form a \textit{unseen instance test set}.
We also randomly split the rest of the collected data (totalling 17,872 samples) into a training set of 12908 samples, a validation set of 3227 samples, and a \textit{seen instance test set} of 1747 samples. Each object (except Chairs 3 and 4) and each action type appear in all three splits.

%% file: pages/Experiments.tex
\section{Experiments}

Comparative evaluation is conducted to validate our network design and the use of object dynamic descriptors. We also examine the model generalizability to unseen objects from the six categories.

\textbf{Training. }
We used the Adam optimizer~\cite{kingma2014adam} with a learning rate of 0.001 for training our network. A mini-batch of 32 samples was fed to the network during training. We trained the neural network models until they converged and showed the best performance on the validation set. All models were trained in a category-agnostic manner across all training data. Training and evaluation were conducted on a machine running Ubuntu 18.04 with a 1.80GHz Intel Xeon Silver 4108 CPU, and an NVIDIA RTX 2080Ti GPU.

\textbf{Evaluation metrics. }
Given the ground-truth (GT) 6D pose changes of an object and the predictions produced by the methods, we measured the errors in pose translation ($mm$) and pose rotation ($10^{-3} rad$). We also adopted the mean per joint position error ($mm$) with respect to the human skeleton (MPJPE-H) and the object keypoints (MPJPE-O). 

\textbf{Ablation setting. }
To justify our design choices, we also evaluate the performance of our network without using the proposed object dynamic descriptors (denoted \textit{Ours w/o Desc}). Specifically, we remove the object dynamic descriptors from the input to the object pre-processing branch and feed only the keypoint information to the weighted pooling block in Figure~\ref{fig:hogcn layout}. The channel size of the FC layer predicting the object motion is adapted accordingly as well.

We further remove all the object information (both its dynamic descriptors and the geometry) and use solely the human skeleton motion in the first 10 frames to predict both the human and object motions in the last 10 frames, which we denoted as \textit{BaseGCN}. In this setting, the input object representation, object encoder, and the weighted pooling block are deducted from our network.

\textbf{Comparison to SOTA methods. }
We compare our model against two state-of-the-art (SOTA) methods (C-TE~\cite{butepage2017deep} and CAHMP~\cite{corona2020context}). All of the methods are given a 1-second observation (10 frames) of the human-object interaction, and required to predict the future 1-second (10 frames) joint motion of the human and the object. We are specifically interested in the prediction of object pose changes as accurate prediction can lead to desirable robot assistive operation in human-robot collaboration tasks.

\begin{table*}[]
    \centering
    \resizebox{0.8\textwidth}{!}{
    \begin{tabular}{cc|cccccccccc|ccc}
        \hline
          &   &   &   &   &   &   &   &   &  &   &  & Short-term & Long-term &  \\
        Metrics & Methods & 0.1s & 0.2s & 0.3s & 0.4s & 0.5s & 0.6s & 0.7s & 0.8s & 0.9s & 1.0s & (0.1s-0.5s) & (0.6s-1.0s) &  Mean  \\
        \hline
        \hline
        Trans. Err. & C-TE & 33.4 & 40.3 & 54.1 & 73.1 & 90.3 & 96.9 & 120.0 & 142.9 & 168.9 & 196.8 & 58.2 & 145.1 & 101.7 \\
        Trans. Err. & BaseGCN & 34.7 & 38.6 & 48.3 & 65.1 & 67.1 & 92.4 & 106.9 & 145.6 & 164.6 & 184.4 & 50.8 & 138.8 & 94.8 \\
        Trans. Err. & Ours w/o Desc & 37.4 & 43.2 & 53.8 & 68.0 & 79.0 & 94.0 & 105.9 & 129.2 & 137.6 & 145.9 & 56.3 & 122.5 & 89.4 \\
        Trans. Err. & Ours & 22.4 & 29.0 & 39.4 & 47.2 & 62.1 & 72.5 & 80.6 & 136.3 & 148.8 & 163.1 & \textbf{40.0} & \textbf{120.3} & \textbf{80.2} \\
        \hline
        Rot. Err. & C-TE & 46.1 & 64.9 & 84.4 & 111.8 & 159.7 & 150.8  & 210.7 & 208.1 & 229.4 & 251.0 & 93.4 & 210.0 & 151.7 \\
        Rot. Err. & BaseGCN & 46.5 & 54.8 & 64.2 & 85.5 & 115.6 & 130.4 & 159.9 & 190.9 & 198.4 & 227.9 & 73.3 & 181.5 & 127.4 \\
        Rot. Err. & Ours w/o Desc & 52.2 & 63.1 & 94.0 & 108.2 & 122.6 & 140.2 & 154.5 & 178.3 & 190.8 & 202.2 & 88.0 & 173.2 & 130.6 \\
        Rot. Err. & Ours & 33.7 & 46.4 & 60.7 & 82.8 & 86.0 & 100.5 & 115.7 & 191.0 & 204.9 & 223.8 & \textbf{61.9} & \textbf{167.2} & \textbf{114.6} \\
        \hline
        \hline
        MPJPE-O & C-TE & 36.4 & 44.7 & 60.1 & 81.5 & 102.1 & 107.3 & 135.3 & 158.0 & 183.4 & 212.4 & 65.0 & 159.3 & 112.1 \\
        MPJPE-O & CAHMP & 25.8 & 42.7 & 59.5 & 75.9 & 92.2 & 108.3 & 124.0 & 139.6 & 155.3 & 171.2 & 59.2 & 139.7 & 99.5 \\
        MPJPE-O & BaseGCN & 37.5 & 42.0 & 53.0 & 71.2 & 76.6 & 101.7 & 117.8 & 157.9 & 175.5 & 198.3 & 56.1 & 150.2 & 103.2\\
        MPJPE-O & Ours w/o Desc & 42.0 & 47.1 & 60.1 & 74.5 & 86.1 & 101.8 & 114.2 & 138.0 & 147.1 & 156.1 & 62.0 & 131.5 & 96.7 \\
        MPJPE-O & Ours & 24.0 & 31.6 & 42.5 & 52.5 & 67.2 & 78.3 & 88.0 & 147.2 & 160.5 & 175.9 & \textbf{43.6} & \textbf{130.0} & \textbf{86.8} \\
        \hline
        MPJPE-H & C-TE & 25.9 & 48.1 & 70.9 & 89.1 & 110.8 & 124.0 & 137.7 & 151.0 & 166.5 & 180.9 & 69.0 & 152.0 & 112.1 \\
        MPJPE-H & CAHMP & 22.3 & 42.4 & 60.0 & 76.0 & 91.1 & 105.7 & 120.1 & 134.7 & 149.7 & 165.1 & \textbf{58.3} & 135.1 & \textbf{96.7} \\
        MPJPE-H & BaseGCN & 27.8 & 50.8 & 73.4 & 92.0 & 108.1 & 123.2 & 138.1 & 143.0 & 159.0 & 166.2 & 70.4 & 145.9 & 108.2 \\
        MPJPE-H & Ours w/o Desc & 24.0 & 45.9 & 65.7 & 82.6 & 97.3 & 110.2 & 122.5 & 134.1 & 144.4 & 154.5 & 63.1 & \textbf{133.2} & 98.2 \\
        MPJPE-H & Ours & 23.9 & 45.8 & 65.8 & 84.6 & 100.6 & 115.4 & 129.4 & 140.0 & 151.9 & 163.6 & 64.1 & 140.1 & 102.1 \\
        \hline
    \end{tabular}}
    \caption{Comparison of performances of different methods on seen objects. The upper block reports the prediction errors regarding the 6DOF pose (i.e., translation and rotation). The bottom block reports the MPJPE with respect to the object and the human.}
    \label{tab:comparison_test}
\end{table*}

\subsection{Quantitative analysis}

Firstly, we evaluate our proposed method (HO-GCN), its variants (Ours w/o Desc and BaseGCN), and the comparing methods (C-TE and CAHMP) on the test set containing \textit{seen objects}. Different methods are compared in terms of each future frame, short-term ($0.1{-}0.5s$) and long-term ($0.6{-}1.0 s$) in Tab.~\ref{tab:comparison_test}. 

Comparisons show that our method consistently outperforms C-TE~\cite{butepage2017deep} which uses a convolution-based encoder-decoder structure for human-object joint motion prediction by a large margin.
While the comparison with CAHMP shows it performs slightly better than our method in terms of human motion prediction, our method achieves a marginal improvement over CAHMP on the MPJPE-O metric in both short term and long term. 

\begin{figure}[h]
\begin{center}
   \includegraphics[width=\linewidth]{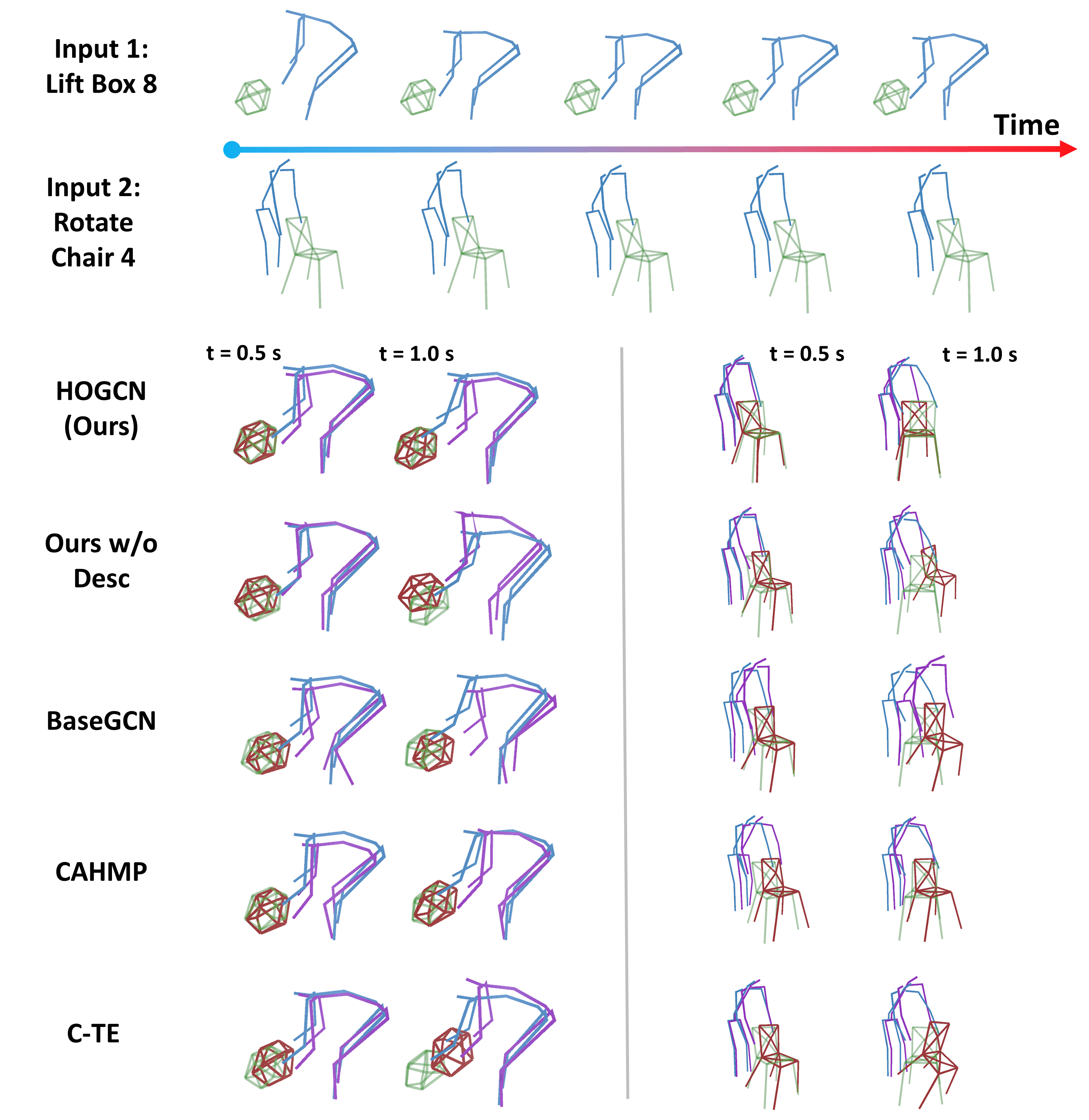}
\end{center}
\caption{Qualitative comparison of different methods. \textit{Lift Box 8} is sampled from the seen object test set while \textit{Rotate Chair 4} is from the unseen one. Results of the human and the object predicted by our HO-GCN are shown in purple and red, respectively, while GT's are drawn in blue and green. Our method can well capture the intended motion.}
\label{fig:qualitative_results_generalize}
\end{figure}

Secondly, we also test the different models on \textit{two unseen objects}. In particular, Chair 3 (see Fig.~\ref{fig:objects}) is an armchair, which is largely different from the other chairs in appearance and geometry. Note that the same keypoint configuration is used as shown in Fig.~\ref{fig:objects} for motion capture. Even without attaching markers to the armrests, we can achieve satisfactory results (see Fig.~\ref{fig:qualitative_results_generalize}), showing that the abstracted geometry is sufficient for predicting the motion of a rigid body in our framework, justifying the use of 12 keypoints for all objects. 

Tab.~\ref{tab:unseen_objects} shows the averaged performances over the 10 future frames. We can see that generalizing the trained models to unseen instances is challenging; compared to their performances in the all-seen setting, all models obtained worse results. 
Under this condition, our HO-GCN outperforms its variant (Ours w/o Desc and BaseGCN) by a relatively large margin on both chairs, showing the efficacy of the object dynamic descriptors for model generalization. 

When comparing to CAHMP~\cite{corona2020context}, we see that our method consistently achieves better results on different unseen chairs regarding the object motion prediction, showing the feasibility of applying our method to human-robot collaborative tasks, while the gap between Ours and CAHMP regarding the human motion prediction is narrowed.  
Both our method based on GCN and CAHMP based on RNN gain large improvement when compared to C-TE~\cite{butepage2017deep}.

\textbf{Will the object dynamic descriptors benefit the task?}
Comparing our method to its variant (Ours w/o Desc) for ablation, we observe consistent performance gains regarding MPJPE-O, the translation and rotation errors in both short/long-term settings. 
From Tab.~\ref{tab:comparison_test}, the gains for the \textit{Seen Objects} test set are observed substantial in the short-term phase, where the motion state of the object is changing as it was just moved by the human actor. Consistent gains are observed in Tab.~\ref{tab:unseen_objects} for \textit{Unseen Objects} test set too.
Such reliable prediction is crucial for predicting the intents of how the human actor is going to manipulate the objects, and supports the use of our object dynamic descriptors for describing the intrinsic physical properties of an object. 

\textbf{Can human skeleton data alone be used to predict the joint motion?}
We also compare the variant of our method (Ours w/o Desc) to the baseline model (BaseGCN) that uses only the human skeleton data to predict the human-object joint motion. We see a small improvement obtained by our variant in the long-term. This may be attributed to that in the long-term the object motion is relatively large, and thus the benefit of using the object information (in this case the object keypoint positions) is more obvious.

\begin{table}[]
    \centering
    \resizebox{\linewidth}{!}{%
    \begin{tabular}{c | c c c c | c}
        \hline
        Method & Tran. Err. & Rot. Err. & MPJPE-O & MPJPE-H & Object \\
        \hline
        \hline
        C-TE & 148.9 & 332.4 & 188.1 & 134.8 & \\
        CAHMP & - & - & 168.9 & 124.5 & \\
        BaseGCN & 135.3 & 311.7 & 168.2 & 131.7 & Chair 3 \\
        Ours w/o Desc & 129.6 & 304.1 & 160.8 & \textbf{124.0} & \\
        Ours & \textbf{122.1} & \textbf{296.5} & \textbf{151.0} & 128.9 &\\
        \hline
        \hline
        C-TE & 175.7 & 394.3 & 214.8 & 108.1 &  \\
        CAHMP & - & - & 165.4 & \textbf{85.8} &  \\
        BaseGCN & 160.4 & 350.2 & 196.2 & 100.9 & Chair 4 \\
        Ours w/o Desc & 131.0 & 319.2 & 164.1 & 91.7 & \\
        Ours & \textbf{124.2} & \textbf{310.8} & \textbf{156.7} & 89.0 &  \\
        \hline
        \hline
        C-TE & 160.7 & 359.6 & 199.8 & 123.1 & \\
        CAHMP & - & - & 167.4 & \textbf{107.5} & Sample\\
        BaseGCN & 146.4 & 328.6 & 180.5 & 118.2 &  Mean \\
        Ours w/o Desc & 130.2 & 310.8 & 162.2 & 109.8& \\
        Ours & \textbf{123.0} & \textbf{302.8} & \textbf{153.5} & 111.4 &    \\
        \hline
        \hline
    \end{tabular}}
    \caption{Model generalization test on \textbf{unseen objects} \textit{Chair 3} and \textit{Chair 4}. Specifically, \textit{Chair 3} is an armchair with drastically different geometry from Chairs (1 and 2) for training.}
    \label{tab:unseen_objects}
\end{table}

\begin{figure*}[h]
\begin{center}
   \includegraphics[width=0.9\linewidth]{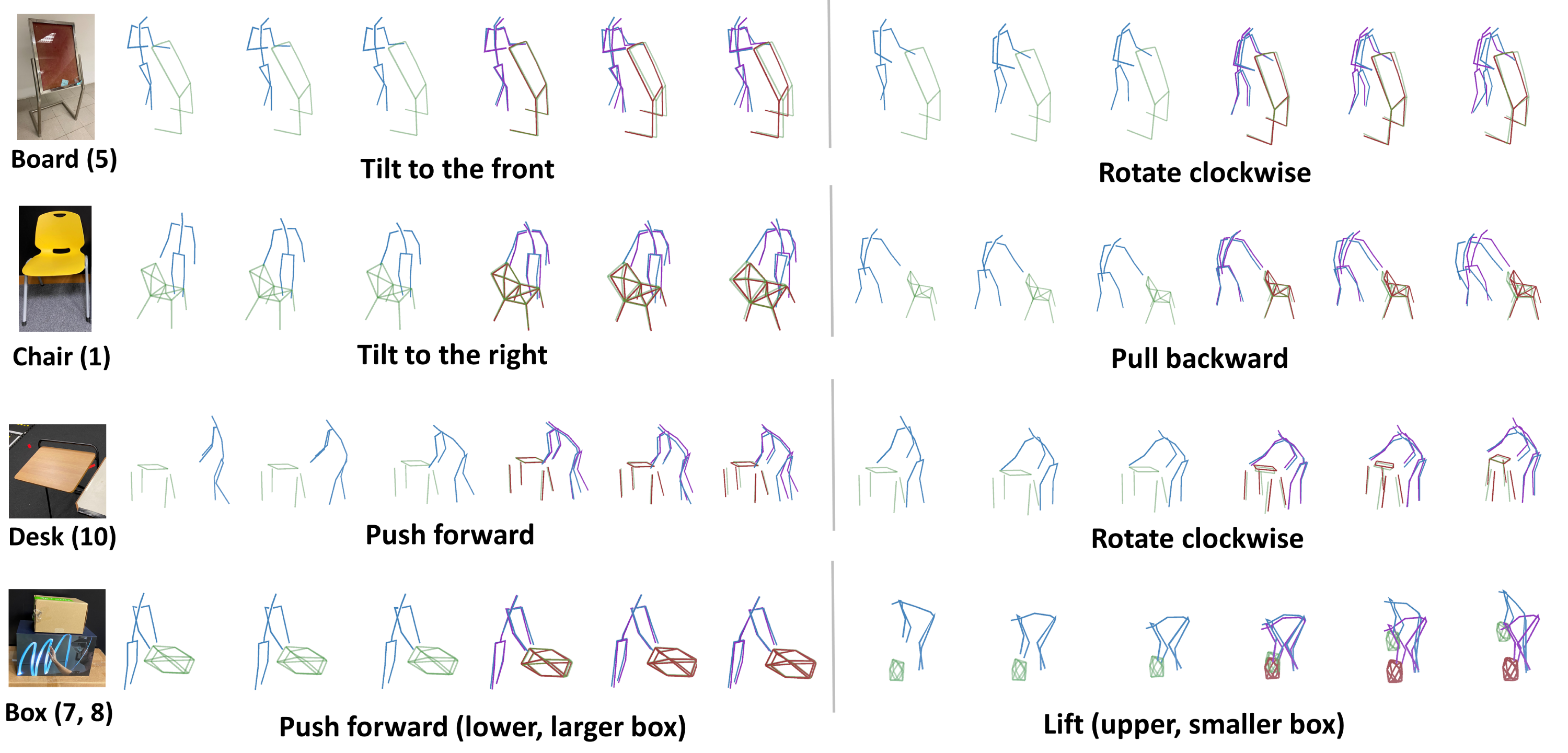}
\end{center}
\caption{Qualitative results produced by HO-GCN with the object dynamic descriptors are shown. Real-world objects are shown on the left. Results of the human and the object predicted by our HO-GCN are shown in purple and red, respectively, while GT is drawn in blue and green. Three frames of the input sequence and three prediction frames are visualized. A failure case is shown at the bottom right.}
\label{fig:qualitative_results}
\end{figure*}

\subsection{Qualitative analysis}

We visualized some randomly sampled results for qualitative analysis and comparison. A collection of qualitative results are shown in Fig.~\ref{fig:qualitative_results}. Objects are displayed on the left. Results of the human and the object predicted by our HO-GCN are shown in purple and red, respectively. We visualize three frames of the input sequence and three prediction frames. 
In this setting, the prediction results are quite close to the GT. We show a failure case (observed across all methods) at the bottom right where the actor was lifting the small box (Object ID 7) and a large difference between GT and the prediction is seen. We assume that because the dataset mainly contains large-size objects such as chairs and tables, the trained model fails to generalize well to this smaller box, while producing a satisfactory result for the medium-sized box (Object ID 8) for the \textit{push forward} action.

Fig.~\ref{fig:qualitative_results_generalize} shows the qualitative comparison of different methods. Two input motion sequences, \textit{Lift Box 8} and \textit{Rotate Chair 4}, are shown at the top. The former sequence was sampled from the all-seen test set while the latter from the unseen. We show two predicted frames at $t=0.5s$ and $1.0s$ corresponding to the short-term and long-term results. All methods produce reasonable results regarding the \textit{Lift Box 8} sequence. However, C-TE and one of our variants (Our w/o Desc) seem to predict motions faster than the ground-truth. For \textit{Rotate Chair 4}, only our method produces a plausible motion of rotating the chair. CAHMP, on the other hand, predicts a rough translation without rotating the chair, failing to capture the intention as the other methods do. 

%% file: pages/demo.tex
\subsection{Human-robot collaborative tasks}
We also showcase that the proposed HO-GCN can be useful for human-robot collaborative tasks. 
Specifically, we are interested in whether HO-GCN can successfully predict the manipulation intent of the human actor in terms of the future motion of the target object. 
We thus conducted experiments with a UR5 robot arm and a vacuum gripper with four silicone suction cups (of diameter 40mm). We chose a box which is suitable in weight and size for demonstration. The end-effector of the robot arm was pre-aligned to the target surface of the box. We used the OptiTrack system to provide 3D human motion information as well as the starting poses of the object.

With an input sequence of a human manipulating the box, our proposed method can predict the potential movement of the object in terms of its 6DoF changes. We then converted it to the corresponding robot trajectory that best produces the predicted object movement.
In this showcase, we didn't consider the leading/following roles of the human and the robot in the cooperative task which could be our future work.

\begin{figure}[h]
\begin{center}
   \includegraphics[width=0.7\linewidth, trim = 150 350 150 0]{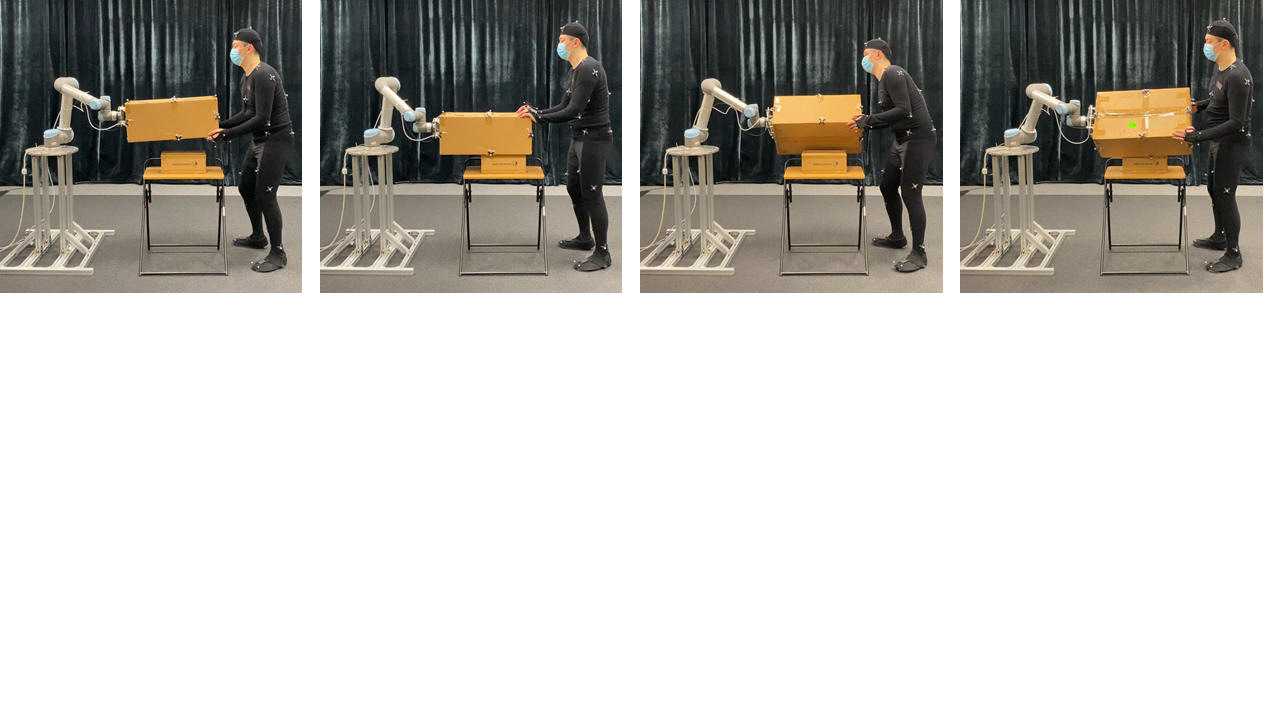}
\end{center}
\caption{Experiment results of the collaborative tasks. The human actors were performing lifting, pushing, clockwise rotating, and counter-clockwise rotating from left to right.}
\label{fig:demo_result}
\end{figure}

Our HO-GCN ran at a real-time rate (42 FPS) for processing such a 10-frame sequence on a desktop running Ubuntu 16.04 with an Intel Core i7-9700K CPU and an NVIDIA RTX 2080 GPU. Some results are shown in Fig.~\ref{fig:demo_result} and more can be found in our supplemental video. 

%% file: pages/conclusion.tex
\section{Conclusions}
In this paper, we focus on predicting full-body human interactions with large-sized daily objects and contribute to a large-scale dataset. Given as input a sequential observation of the human-object interaction, we design a novel graph convolutional network for predicting the future interactive motion. 
We also show that the object dynamic descriptors encoding the inherent physical properties of an object are beneficial to our network's generalization to unseen instances during test.
Some showcasing examples of human-robot collaboration were presented using the proposed motion prediction method.

Currently, a naive version of the dynamic descriptor has been investigated; finding a more powerful representation for the inherent dynamic properties of an object is a promising future work. We would like to extend our human-robot collaboration to more objects and realistic settings. This would require extending our current work to predicting human-object interactions from video inputs.